\documentclass{article}
\usepackage{ijcai20}
\usepackage{xcolor}
\usepackage[normalem]{ulem}
\usepackage{times}

\usepackage{soul}
\usepackage{url}
\usepackage[hidelinks]{hyperref}
\usepackage[utf8]{inputenc}
\usepackage[small]{caption}
\usepackage{graphicx}
\usepackage{amsmath}
\usepackage{booktabs}

\title{2nd Place Solution for IJCAI-PRICAI 2020 3D AI Challenge: 3D Object
Reconstruction from A Single Image}

\author{
Yichen Cao$^1$\and
Yufei Wei$^1$\and
Shichao Liu$^{1}$\and
Lin Xu$^1$\footnote{Contact Author}\\
\affiliations
$^1$Institute of Artificial Intelligence, Shanghai Em-Data Technology Co., Ltd.\\
\emails
\{caoyichen, weiyufei, liushichao, xulin\}@em-data.com.cn
}

\begin{document}

\maketitle

\begin{abstract}


In this paper, we present our solution for the {\it IJCAI--PRICAI--20 3D AI Challenge: 3D Object Reconstruction from A Single Image}. We develop a variant of AtlasNet that consumes single 2D images and generates 3D point clouds through 2D to 3D mapping. To push the performance to the limit and present guidance on crucial implementation choices, we conduct extensive experiments to analyze the influence of decoder design and different settings on the normalization, projection, and sampling methods. Our method achieves the 2nd place in the final track with a score of $70.88$, chamfer distance of $36.87$, and a mean f-score of $59.18$. The source code of our method will be available at \url{https://github.com/em-data/Enhanced_AtlasNet_3DReconstruction}.

\end{abstract}

\section{Introduction}
Significant progress has been made on learning good representations for images, allowing impressive applications in image generation. However, it is still challenging to acquire accurate and complete 3D shapes and scenes due to shape geometry, surface material, and lighting conditions. However, their direct and naive extension from 2D images to 3D voxels introduces high memory costs and inefficient computation issues. For balancing computation and performance, it seems generating point cloud is widely used.

In this paper, we develop a generative neural network which outputs point clouds natively. In this challenge, Chamfer Distance (CD), one of the metrics, can compare two point sets while it does not consider the surface/mesh connectivity. For example, if we have a cube that only has eight vertices, the point cloud generated by our model closer to these eight vertices,
the small Chamfer Distance we will get. Another metric is F-score is also base on Chamfer distance. If we use implicit function models, it may generate high visual quality 3D models, generating a more smooth and continuous surface rather than points close to vertices. As a result, we choose AtlasNet as our baseline, which directly generates point clouds. What is more, In AtlasNet \cite{groueix2018papier} and IM-NET \cite{chen2019learning}, they all trained a 3{D} autoencoder (AE) first and then used it to help the single-view reconstruction training. This approach indicates that it is easy for networks to learn from 3{D} data due to single-view reconstruction, especially an ill-pose problem. Inspired by this, we replace the AtlasNet's decoder to FoldingNet \cite{yang2018foldingnet} decoder. FoldingNet mentioned a novel folding-base decoder deforms a canonical 2D grid onto the underlying 3D object surface of a point cloud, achieving low reconstruction errors even for objects with delicate structures.

\section{Related work}
Object-based single view reconstruction (SVR) calls for generating an object's 3D model given a single image. Recently, there are lots of work in learning and generative modeling 3D shapes.
It is still challenging to acquire accurate and complete 3D shapes due to the interference by shape geometry, surface material, lighting conditions, and the noise introduced in the capturing process.
Up to date, deep neural networks for shape reconstruction might be categorized into three categories: 
voxel-based representations \cite{Choy20163DR2N2AU,liao2018deep}, point-based representations \cite{groueix2018papier,fan2017point}, and implicit fucntions representations \cite{mescheder2019occupancy,chen2020bsp,chen2019learning}.

\noindent\textbf{Voxel representations} are a straightforward generalization of pixels to 3D case. Unfortunately, the memory cost for voxel representations grows cubically with resolution, therefore, it can only generate $32^3$ or $64^3$ voxels. While it's possible to reduce memory cost by using special data structures such as octrees, this structure leads to complex algorithms and is still limited to relatively small $256^3$ voxel grids.

\noindent\textbf{Point-based representations} are widely used both in computer graphics and robotics. Pointnet \cite{qi2017pointnet} pioneered point clouds as a representation for discriminative deep learning tasks. They achieved permutation invariance by applying a global pooling operation.
As a result, a category of techniques pioneered by PointNet that express surfaces as point clouds, and techniques pioneered by Atlas that adopt 2D-to-3D
 mapping process \cite{groueix2018papier}. But these methods do not guarantee watertight results and hardly scale beyond a hundred vertices

 \noindent\textbf{Implicit functions representations} model shapes as learnable indicator functions rather than samples, as in the case of voxel methods. The resulting networks treat reconstruction as a classification problem and are universal approximators whose reconstruction precision is proportional to the network complexity. However, at inference time, these methods still require the execution of an expensive iso-surfacing operation whose performance effected by the desired resolution. Compared with other methods, implicit functions representations can achieve higher visual quality.

\begin{figure*}[h]
    \centering
     \includegraphics[width=\linewidth]{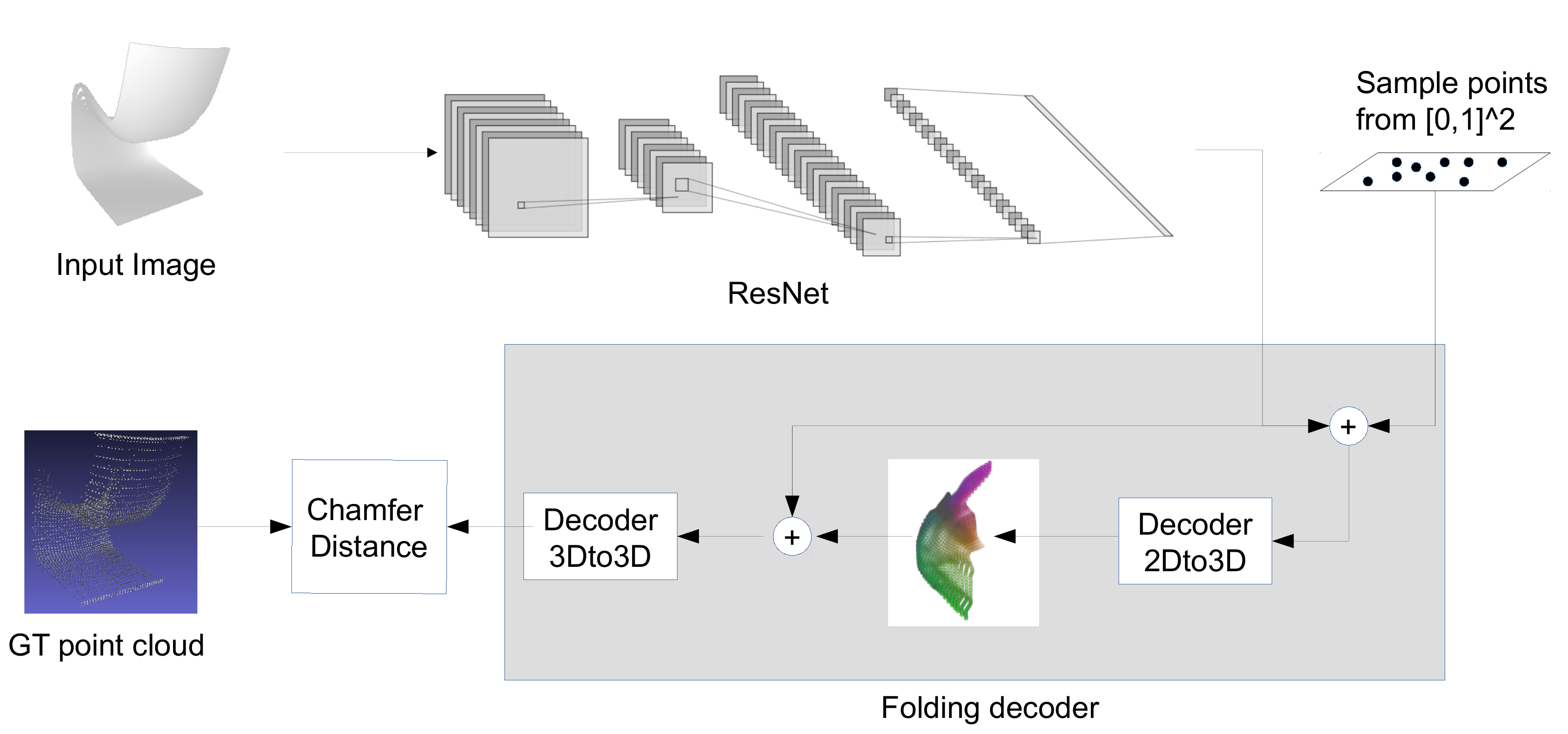}
  
    %
    %
  \caption{\label{fig:arch} \noindent\textbf{Network Architecture. } Given an input image \textit{I}, we employ a ResNet to extract the latent vector.
  The input of $Decoder2Dto3D$ and the $Decoder2Dto3D$ all need to concatenate the output of ResNet. 
  We further optimize the whole network by using Chamfer distance. \textcircled{+} means concatenate}
  \end{figure*}

\section{Method}
\label{sec:model}
\noindent\textbf{Overview. }
Fig.~\ref{fig:arch} shows an overview of our method. For single-view 3D reconstruction,
Actually, we did not follow the idea from AtlaNet to first train an autoencoder by only using point cloud data,
then fix the parameters of the decoder when training SVR.
We replace the original AtlasNet's decoder with FoldingNet's decoder.
In our experiments, we adopted a more radical approach by only training the whole network to minimize the mean CD loss between the predicted and ground truth point clouds.

\noindent\textbf{Backbone. }We used ResNet \cite{he2016deep} as our image encoder to obtain 1024D features from $224 \times 224$ images.
In addition, all relu layers in ResNet were replaced with leakly-relu layers.

\noindent\textbf{Data pre-processing. }For data pre-processing,
since there are multi-view images  and the scales of point clouds are different,
We use the official code to rotate the point cloud to the related image's view and then normalize the point cloud.
After normalization, a trade-off is between performance and computation is down-sampling the original point cloud to 2048 points.
Also, we add a small Gaussian noise to each point to improve the robustness of our network.
For image augmentation, we only keep resizing images to 224, which is different from AtlasNet.

\noindent\textbf{Decoder design.  }To satisfy the submission requirements, the number of points in the predicted point cloud is 2048.
so that the primitives of the decoder are eight, and the number of each decoder's layer is 3.
The input latent vector of the $Decoder2Dto3D$ is obtained from the image encoder.
We replicate it $n$ times and concatenate the latent $n$-by-1024 matrix with an $n$-by-2 matrix that contains the n sampled points from $[0,1]^2$.
The result of the concatenation is a matrix of size $n$-by-514.
The matrix is processed row-wise by a 3-layer perception, and the output is a matrix of size$ n$-by-3 output and feeds it
into a 3-layer perception. For the $Decoder3Dto3D$,  it is the same as the $Decoder2Dto3D$. The only difference is the input of its is an $n$-by-515 matrix.

\noindent\textbf{Sampling method.  }During the training stage, random sample points from $[0,1]^2$ can improve the robustness of the whole network.
AtlasNet noted that sampling points regularly on a grid on the learned
parameterization yields better performance than sampling points randomly. For inference, all results used this regular sampling.

\noindent\textbf{End-to-end training.  } 
We briefly describe how the whole network is built on top of the AtlaNet and FoldingNet for end-to-end training.
Given a 2D input image \textit{I}, our goal is to learn making a prediction for the ground-truth $\emph{S}$. 
Let the prediction of the final reconstruction output be $\widetilde{\emph{S}}$. Both of them are point sets.
The learning process tries to minimize the $Chamfer loss$ between the $\emph{S}$ and $\widetilde{\emph{S}}$:
\begin{equation}
    L(\emph{S}, \widetilde{\emph{S}})={ \frac{1}{|S|}\sum_{x  \in S} \min_{{y} \in \widetilde{S}} \|x-y \|_2}+{ \frac{1}{|\widetilde{S}|}\sum_{y  \in \widetilde{S}} \min_{{x} \in S} \|x-y \|_2}
\end{equation}
the term  $\min_{y \in \widetilde{S}}\|x-y\|$ 
enforces that any 3D point x in the ground-truth has a matching 3D point y in the reconstruction point cloud, 
and the term $\min_{x \in S}\|x-y\|$ enforces the matching vice versa. 
The $\frac{1}{|S|} $ and $\frac{1}{\widetilde{|S|}}$ are all mean operation to avoid too many points cause the huge loss.
In FoldingNet, they used max operation instead of mean operation but if there is an outlier, it will make loss huge and then, and then the next predicted point cloud still has an outlier so repeatedly. It is not easy to make the network converge.

\section{Experiment}
\label{sec:results}
This section shows qualitative and quantitative results on this task using our network and comparing them with AtlasNet.
To compare with the baseline, we first random sample points from the original point cloud to 2048 points. We use the dataset provided by \cite{fu20203dfuture},
We only use 5202 normalized models and their related 62424 multi-view images.
We evaluated our method on trackA and trackB server, which are provided by IJCAI-PRICAI 2020 3D AI Challenge:3D Object Reconstruction from A Single Image.

\noindent\textbf{Evaluation criteria. }
We evaluated our generated shape outputs by comparing them to ground truth shapes using two criteria.
First, we compared point sets of the output and ground-truth
shapes using CD. But CD only reflects the distance between the two point clouds as a whole.
Let the precision of a reconstructed model be the percentage of reconstructed points that have a ground truth point within distance $\tau$. So we introduce F-score to measure the accuracy of each point.
Let the recall of a reconstructed model be the percentage of ground-truth points
that have a reconstructed point withi distance $\tau$. For evaluating above two metrics, the overall score in trackA:
\begin{equation}
  score={100\times\frac{2-CD}{2}\times0.5+{Fscore}\times0.5}
\end{equation}
In trackB:
\begin{equation}
  score=100\times\frac{2-CD}{2}\times0.3+{Fscore}\times0.7
\end{equation}

\noindent\textbf{Ablative analysis. } 
During the competition, several operations on the ground truth point clouds affect the final result.

\noindent\textbf{Normalization method. } In Table  \ref{table:t3}, different normalization methods impact results. 
Sphere normalization means to find the farthest point from the origin and then use its distance from the origin to normalize all points.
In this way, for compact objects such as desk or nightstand, the points on their surfaces are not very far away from a specific point, 
which means all points could keep their relative position. However, for some specific objects like a chandelier, 
it has at least one outlier point, which makes the normalized length is too considerable for the other points. 
After normalization, the whole point cloud could not keep its original shape because of more points closer to the origin point.
Due to the essence of AtlasNet is to learn the mapping function from 2D points to 3D points, 
the network is sensitive to the correspondence between 2D points and 3D points. 
In summary, the key to getting more accurate results is to keep the whole point cloud shape no matter what to do any augmentation.
\begin{table}[h]
    \centering
    
        \begin{tabular}{|c|c|c|c|}
         \hline
              Methods &CD$\downarrow$ & F-score$\uparrow$ &score$\uparrow$ \\
         \hline
    
           AtlasNet(baseline) & 4.2 &84.97  &  91.43\\
         \hline
           SQ & 1.49 & 96.59 & 97.92\\

         \hline
       \end{tabular}
       \caption{Reconstruction metrics comparison on trackA. 
       baseline: use sphere normalization
       SQ: use square normalization, default is unitball normalization;
       Chamfer distance is reported, multiplied by $10^{2}$}  
       \label{table:t3}
    \end{table}

    \noindent\textbf{Projection. }In Table \ref{table:t7}, whether to use multi-view images to rotate original point clouds or not could bring a huge difference.
    In Table \ref{table:t4},  different scale factors lead to distinguishing results.
    Using multi-view images to rotate original point clouds is similar to calculate the relative projection matrix.
    Using different scale factors is like to guess the focal length.
    To sum up, it is not only lead by the reason from Table \ref{table:t3} but also relative to the projection transformation. 
    If the camera matrix and projection matrix are known,  we can reconstruct point clouds from single RGB images. 
    As a result, our network also estimated the camera matrix and projection matrix implicitly.

\begin{figure*}[h]
        \centering
         \includegraphics[width=\linewidth]{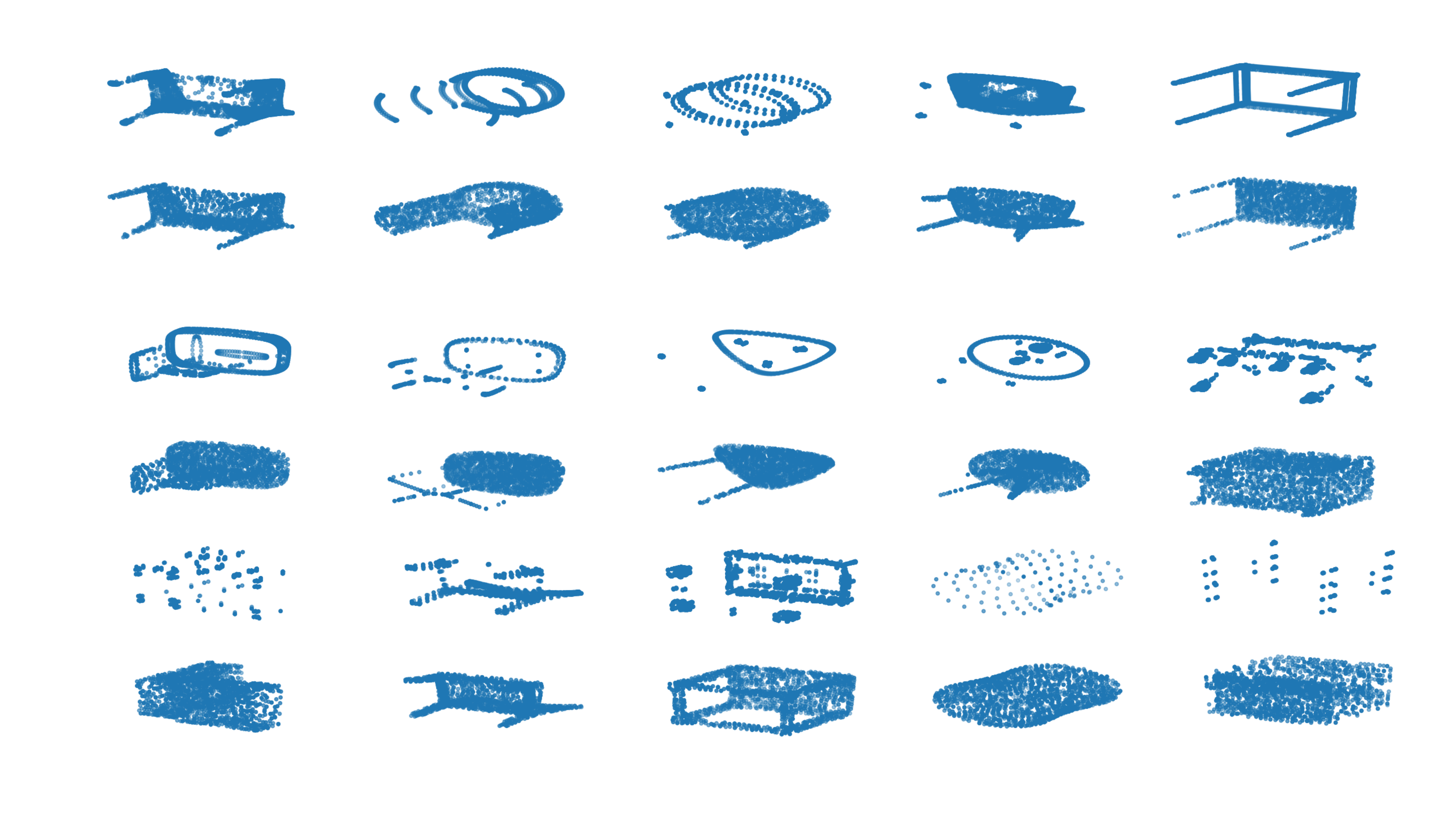}
      
      \caption{\label{fig:sample} Sample results, odd rows are original point cloud, even rows are point cloud after sampling mesh with Lloyd's algorithm. }
      \end{figure*}

    \noindent\textbf{Sampling method. } 
    Fig. \ref{fig:sample} shows the point cloud before and after sampling mesh with lloyd's algorithm\cite{sample}.
    After sampling, the density of point clouds increases. For the whole point cloud, their surfaces are more continuous. 
    It seems that it's not difficult to learn the mapping function. 
    However, Table \ref{table:t8} shows that the network trained by the sampled point cloud is not better than the network trained by the original point cloud.
    In this way, we found that Chamfer distance is sensitive to the location of the given points. 
    In this challenge, the points in the ground-truth point cloud are all vertices, 
    which hopes all predicted points could be close to the edge or corner rather than distribute continuous surfaces.
    For visualization purposes, the more point clouds distribute on the surface, the more comfortable people observe. 
    As a result, Chamfer distance maybe not enough to measure the quality of the generated point cloud. 
    It needs to introduce some criteria for visual quality or generated mesh.

      \begin{table}[h]
        \centering
        
            \begin{tabular}{|c|c|c|c|}
             \hline
                  Methods &CD$\downarrow$ & F-score$\uparrow$ &score$\uparrow$ \\
             \hline
               AtlasNet(w/o sampling) & 1.45 &96.85  &  98.06\\
             \hline
             AtlasNet(w sampling & 1.49 &85.06  &  97.92\\
             \hline
           \end{tabular}
           \caption{Reconstruction metrics comparison on trackA. MV: use multi-view image to rotate original point clouds;
           Chamfer distance is reported, multiplied by $10^{2}$}  
           \label{table:t8}
        \end{table}
    \begin{table}[h]
        \centering
        
            \begin{tabular}{|c|c|c|c|}
             \hline
                  Methods &CD$\downarrow$ & F-score$\uparrow$ &score$\uparrow$ \\
             \hline
               AtlasNet(baseline w/o MV) & 16.82 &52.92  &  72.26\\
             \hline
             AtlasNet(baseline w MV) & 4.2 &84.97  &  91.43\\
             \hline
           \end{tabular}
           \caption{Reconstruction metrics comparison on trackA. MV: use multi-view image to rotate original point clouds;
           Chamfer distance is reported, multiplied by $10^{2}$}  
           \label{table:t7}
        \end{table}

    \begin{table}[h]
        \centering
        
            \begin{tabular}{|c|c|c|c|}
             \hline
                  Methods &CD$\downarrow$ & F-score$\uparrow$ &score$\uparrow$ \\
             \hline
        
               $\times0.001$ & 4.1 &85.31  &  91.63\\
               $\times0.01$ & 2.53 &91.52  &  95.13\\
               $\times0.02$ & 1.47 &96.85  &  98.06\\
               $\times0.1$ & 1.43 &96.79  &  98.03\\
               $\times1$ & 1.46 &96.75  &  98.01\\
               $\times10$ & 1.45 &96.84  &  98.06\\
               $\times50$ & 1.43 &96.97  &  98.13\\
               $\times100$ & 1.41 &96.98  &  98.08\\
             \hline
           \end{tabular}
           \caption{Reconstruction metrics comparison on trackA. 
           $\times$n: means to scale the original point cloud by n, which means any point in ground-truth multiples n.
           Chamfer distance is reported, multiplied by $10^{2}$}  
           \label{table:t4}
        \end{table}

\noindent\textbf{Evaluation on trackA server. }
We report quantitative results for shape generation from single images in Table \ref{table:t5},
where each approach is trained on all categories and results are averaged over all categories.
According to the ablative analysis, we combine all tricks to get the best results. 

\begin{table}[h]
    \centering
    
        \begin{tabular}{|c|c|c|c|}
         \hline
              Methods &CD$\downarrow$ & F-score$\uparrow$ &score$\uparrow$ \\
         \hline
           AtlasNet(baseline w MV) & 4.2 &84.97  &  91.43\\
         \hline
           Our+SQ & 1.49 & 96.59 & 97.92\\
           \hline
           Our+SQ+FD  &1.45&96.85&98.06\\
           \hline
           Our+SQ+FD+S  &\textbf{1.41} &\textbf{96.98}&\textbf{98.13}\\
         \hline
       \end{tabular}
       \caption{Reconstruction metrics comparison on trackA. SQ: use square normalization, default is unitball normalization;
       FD: use decoder from FoldingNet; S: scale original point clouds, in our experiment, we use 50.
       Chamfer distance is reported, multiplied by $10^{2}$}
       \label{table:t5}
\end{table}

\section{Conclusion}
\label{sec:conclusion}
We have introduced our approach to the reconstruction point cloud from a single image.  We have shown its benefits for 3{D} single-view reconstruction, out-performing existing baselines. Specifically, we use both a single-view reconstruction model and a 3{D} auto-encoder to yield robust and accurate reconstruction.
We also employ many empirical settings on the normalization, projection, and sampling trials to boost 
performance.
Consequently, our proposed method achieved the 2nd place in the {\it IJCAI--PRICAI--20 3D AI Challenge: 3D Object Reconstruction from A Single Image}.

\noindent\textbf{Acknowledgements } Our work is supported by Em-Data Technology Co., Ltd. 
We thank Jianfei Gao, Fengliang Qi, Yuan Gao, Changyong Shu, Yunjie Xu, Qi Liu, Yao Xiao for valuable discussions.

\bibliographystyle{named}
\bibliography{ijcai20}
\end{document}